# Quantum Inspired Chaotic Salp Swarm Optimization for Dynamic Optimization

Sanjai Pathak [1]
*Amity University Uttar Pradesh*
Noida, India
pathak.sanjai@gmail.com

Ashish Mani [2]
*Amity University Uttar Pradesh*
Noida, India
amani@amity.edu

Mayank Sharma [3]
*Amity University Uttar Pradesh*
Noida, India
msharma22@amity.edu

Amlan Chatterjee [4]
*California State University*
Dominguez Hills Carson, CA, USA
achatterjee@csudh.edu

*Abstract –* **Many real-world problems are dynamic optimization problems that are unknown beforehand. In practice, unpredictable events such as the arrival of new jobs, due date changes, and reservation cancellations, changes in parameters or constraints make the search environment dynamic. Many algorithms are designed to deal with stationary optimization problems, but these algorithms do not face dynamic optimization problems or manage them correctly. Although some optimization algorithms are proposed to deal with the changes in dynamic environments differently, there are still areas of improvement in existing algorithms due to limitations or drawbacks, especially in terms of locating and following the previously identified optima. With this in mind, we studied a variant of SSA known as QSSO, which integrates the principles of quantum computing. An attempt is made to improve the overall performance of standard SSA to deal with the dynamic environment effectively by locating and tracking the global optima for DOPs. This work is an extension of the proposed new algorithm QSSO, known as the Quantum-inspired Chaotic Salp Swarm Optimization (QCSSO) Algorithm, which details the various approaches considered while solving DOPs. A chaotic operator is employed with quantum computing to respond to change and guarantee to increase individual searchability by improving population diversity and the speed at which the algorithm converges. We experimented by evaluating QCSSO on a well-known generalized dynamic benchmark problem (GDBG) provided for CEC 2009, followed by a comparative numerical study with well-regarded algorithms. As promised, the introduced QCSSO is discovered as the rival algorithm for DOPs.**

*Nomenclature:*

| | |
|---|---|
| $w_t$ | value of the chaotic map at the t-iteration |
| $z_n$ | part of a generic superposition for n-qubits |
| $X^i$ | Position of $i^{th}$ Salp |
| $x_j^i$ | position of $i^{th}$ Salp in $j^{th}$ dimension |
| $A_j$ | local attractor for the convergence speed in the search space $j^{th}$ dimension |
| $B_l$ | known as contraction-expansion coefficient in $l^{th}$ generation |
| $u_d$ | chaotic operator equation in $d^{th}$ dimension |
| $F_n$ | $n^{th}$ benchmark function |
| $T_n$ | $n^{th}$ change type |

## I. INTRODUCTION

Meta-heuristic methods are popular for solving complex optimization problems, an intelligent approach in which an iterative process enhances the obtained solution until a concluding state is achieved. Most meta-heuristic methods are designed and implemented to work on static optimization problems, where the search space and the problem environment remain unchanged. The algorithm strived to achieve global optima during optimization [1]. But, most real-world optimization problems nowadays are dynamic and stochastic optimization problems, where the problem of domain space changes throughout the optimization process, and the obtained solution is not more relevant after environment changes. For example, the gantry crane scheduling task is a problem that is generally determined as a stationary optimization. But, entering another job throughout the planning process or the failure of a gantry crane, the search area environment changes from static to dynamic. The previously obtained solution may no longer apply to the new problem space. This kind of problem is known as a dynamic optimization problem in works of literature [2][3].

The discovery of global optimization is the primary objective while solving stationary optimization problems, but finding only global optimal is insufficient for dynamic optimization problems. However, identifying and tracking global optimal is critical for DOPs in dynamic environments. The optimization methods designed for static optimization problems do not follow the optimal appropriately. Therefore, such techniques are not appropriate for dynamic optimization problems. It is essential to find the different methods that align with the objective of dynamic optimization problems and other assessment benchmarks for optimization in dynamic and uncertain environments.

The DOP solution (i.e., defined as S), which constantly evolves through the optimization method, influences the execution of many real-world applications [4].

$$S = z\,(y,\, \emptyset,\, t)$$

when $S$ must be the dynamic optimization problem specified as cost function $z$ inclusive $y$ as a feasible solution from the set of solution $Y$, $\emptyset$ is the control parameter to find the solution distribution in the fitness landscape, and $t$ is the time. The search strategy of the algorithms must be efficient to localize and follow the evolution of the global optima in time $t$ toward finding superior solutions in the fitness landscape for DOPs. Several techniques was suggested in the bibliography to deal with dynamic optimization problems. Meta-heuristic methods were used frequently, including swarm intelligence. The schemes which are used with meta-heuristic methods for the dynamic optimization problems are diversity schemes [5] [6], memory schemes [7] [8], multi-population schemes [9] [10], adaptive schemes, [11] [12], multiobjective optimization for dynamic environments [13] and an adaptive quantum-inspired evolutionary algorithm (AQiEA) for optimizing power losses by dynamic load allocation on distributed generators [42].

Particle Swarm Optimization (PSO) is an approach of optimization based on swarm intelligence, a population-based stochastic optimization approach. PSO is inspired by the social behavior of flocking birds and the schooling of fish. Like other evolutionary algorithms, PSO has outperformed in solving many real-world static optimization problems [1]. However, DOPs are difficult for the standard PSO to solve due to outdated memory when the environment change and diversity loss. A clustering PSO (CPSO) is introduced to address the critical issues of PSO for DOPs, with a local search strategy and clustering method for locating and following multiple optima in the dynamic environment [14]. Many researchers have reported the efficiency of quantum-inspired evolutionary algorithms (QiEA) for solving several combinatorial and benchmark problems in a variety of areas, including Engineering optimization problems [28], Data clustering [36], and Image Processing [37]. In a similar thought, we attempt to extend the standard SSA for the DOPs known as QSSO [10], which is based on a set of methods, including quantum computing, multi-population, and an intelligent shifting operator, to effectively explore the search space during the optimization process in dynamic environments. Further, this paper details the approaches considered while solving DOPs [10]. The generalized dynamic benchmark problem (GDBG) provided for CEC'09 has been employed to assess the presented QCSSO, a well-known standard benchmark problem for evaluating optimization algorithms in dynamic environments.

Salp Swarm Algorithm (SSA) was introduced by Mirjalili et al. in 2017. SSA is a population-based metaheuristic optimization technique that impersonates the swarming behavior of Salp in the ocean by establishing a Salp chain [15]. Several conducted research mainly related to the advancement of SSA for solving real-world problems. A simple SSA with a random search radius was submitted to improve the proficiency of SSA [31]. A particle-based approach for SSA with global exploring and local exploiting was introduced for convergence speed and accuracy [32]. Hybrid SSA with a gravitational search algorithm was studied in [33] to boost its searchability. An elite-based SSA is introduced for numerical optimization problems by improving the searchability of the algorithm [43].

Further, SSA is widely employed in various engineering fields, such as controller placement problems [34] and multilevel color image segmentation [35]. All these applications have demonstrated the pertinency and efficiency of the SSA. The SSA is similar to other evolutionary algorithms in many characteristics and works effectively for many real-world applications. The swarming behavior of Salp in SSA can avoid converging each solution into a local optimum up to a few degrees due to the Salp chain [16]. But there are many optimization problems in the real world where it is difficult for the standard SSA to work efficiently, and sometimes it fails to optimize. Dynamic optimization problems (DOPs) are another domain where optimum global changes over time and SSA fails to improve the obtained global best solution to accomplish the expected global optima in the dynamic space space. The problem lies in SSA for DOPs primarily because of not having a good search scheme and loss of population diversity, when this must boost the global best solution obtained so far to achieve the expected global optimum. The standard SSA search strategy was designed to achieve global optima for static optimization problems. It cannot work as expected for problems in dynamic and uncertain environments.

For DOPs, SSA needs to improve with a good search strategy where it is required to enhance the obtained global best solution and improve the population diversity to prevent stagnation in local optima. A multi-population mechanism in QCSSO is employed to locate and track the multiple local optima throughout the optimization procedure. Further, this paper examines and compares the performance of QCSSO with well-regarded algorithms QSSA, Standard Particle Swarm Optimization (SPSO), and Clustering Particle Swarm Optimization (CPSO). However, SSA has been evaluated in several applications, especially on engineering design optimization problems, but hardly ever employed on DOPs, as seen from the survey [17].

The paper is structured as follows: Section II presents the relevant related work. Section III presents the proposed algorithm QCSSO with the techniques used for DOP, and section IV presents the experiment configurations. Experimental evaluation and discussion are presented in section V. Finally, the conclusion and relevant future work are presented in section VI.

## II. RELATED WORK

There are specific challenges for the meta-heuristic methods in dynamic environments during the optimization process that is not there for static optimization problems: (1) A good search strategy for DOPs to locate and track global optimum, (2) Outdated memory, (3) Diversity loss.

In the dynamic environment, the fitness of the attained solutions changes due to the dynamic environments. It will no anymore coincide with the retained value in the memory exploited by the algorithm. Loss of diversity occurs in every meta-heuristic method due to an intrinsic nature of convergence to the global optima. Generally, the loss of diversity reason is the built-in property of meta-heuristic algorithms for convergence to the previous optimal position and the excessive vicinity of the solutions to each other.

The re-initialization is a straightforward way to prevent convergence on the previous optimum stance and excessive closeness of the solutions[18]. In re-initialization, the method considers the changing environment as a new optimization problem and relaunches the optimization process using the evolved environment. But the efficiency of the optimization algorithm could improve upon the acquired knowledge of the earlier environment, and the re-initialization technique insinuated the depletion of all obtained knowledge so far from the search space. Although outdated memory is of less relevance in comparison to diversity loss, there are solutions proposed in the literature to handle it [19]: (1) Forgetting memory, (2) Re-evaluating memory. These two proposed solutions apply to the optimization approaches where the obtained information from the search space is stored. In the forgetting memory technique, the position preserved for each solution will be substituted by the positions of the new environment. In the re-evaluating memory technique, the saved positions in memory are re-evaluated.

The diversity loss is because of the built-in character of the meta-heuristic technique, designed originally for static optimization, where quick convergence has been considered a good feature. The works of literature proposed several solutions, including a memory-based approach, to store previous optimal solutions to use when the environment changes, mutation, and self-adaption in which diversity loss is allowed and later to solve it for the expected outcome of the optimization algorithm. An adaptive mutation operator, i.e., activated Hyper-Mutation, was suggested as a factor to be multiplied by the specific mutation to create diversity [20]. In [21], an adaptive approach is employed for the chaotic mutation to create diversity in the environment. Another technique was introduced as a local variable search to address the consistency problem in the mutation step size by making it adaptive. In [23], replacing random solutions with formerly obtained solutions in case of environmental changes is an alternative suggestion proposed to produce diversity. In [24], an implicit memory technique presents diploid genetic algorithms for dynamic optimization.

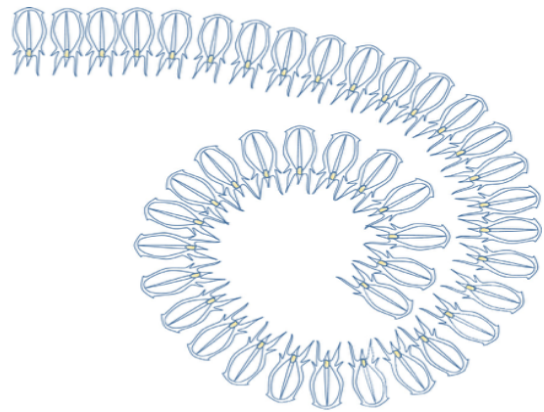

Fig. 1. The Salp Chain.

Multi-population is another approach in uncertain and dynamic environments, considered a hybrid model to create and maintain diversity for the DOPs. In this technique, the population is divided into subpopulations, covering various search space regions. In general, all the subpopulations have similar tasks but can also have distinct tasks. In [25], an approach based on a genetic algorithm was proposed as the Shifting Balance Genetic Algorithm (SBGA). In SBGA, many subpopulations are in charge of global search, and a large subpopulation monitors the evolving peaks. In [26], another approach based on self-organizing scouts is proposed, termed Self-Organizing Scouts (SOS). The SOS conversely utilizes a large subgroup for global search and a few subpopulations for monitoring changes. This policy has also been suggested with other optimization algorithms, such as Genetic Algorithm [27]. In [29] a procedure for optimization in dynamic environments is presented as compound particles. This procedure has an agreeable throughput, and in [30] a new method-built artificial fish swarm algorithm to show a similar approach.

Like any other evolutionary algorithm, quick convergence is a good feature in SSA and benefits many static optimization problems. The fast convergence is mainly due to the loss of diversity in the population, and control parameters are fixed during the optimization process. In the SSA, all the Salp attracted strongly towards food position and sometimes converged on local or global optima where food position is located. However, because of this feature, the original SSA considers the optimization problem a single instance problem and does not adapt when the environment changes. The result is unsuitable for DOPs, where a change reflects an entirely new problem, in which locating and tracking the evolving global optima in the search landscape is necessary. Standard SSA needs to ameliorate by adapting the different mechanisms to control diversity. Controlling diversity is an excellent strategy for DOPs that support converting good local optima into global best in the changing environment during optimization.

## III.   The Quantum-inspired Chaotic Salp Swarm Optimization Algorithm

A good exploring and exploitation tendency of the SSA heuristic on static optimization problems enables it to attract the DOPs. The standard SSA has the superiority of specific updating functions but still has the problem of quick convergence due to loss of diversity and readily falls into the local optimum, particularly when encountering multidimensional and dynamic optimization. Motivated by the thought of implementing the quantum computing technique for metaheuristics to stimulate their global optimization achievement, this paper proposes Quantum Inspired Chaotic Salp Swarm Optimization (QCSSO) Algorithm with a multi-population mechanism to locate and track the optima. Implementing a chaotic series based on a chaotic logistic map in place of random series is a sound approach to diversifying the population and enhancing the QCSSO execution in prevention from early convergence to local optima. The fundamentals of quantum computing and the detailed QCSSO process are developed in this section.

### A.   Quantum Computing

Quantum computing (QC) is an emergent and novel approach to computational intelligence. It is advanced in the concepts and principles of quantum physics. This integration emerged from the knowledge of quantum computers, where specific calculations are carried out more quickly than digital computers. An accelerated analysis is made possible using the quantum principles of computation, such as the superposition of states, entanglement, and interference [36][37]. In a quantum computer, a particle can be in a superposition state where two or more quantum states can be added together, resulting in another valid quantum state. Unlike bit (0 or 1) in a digital computer, a quantum bit or qubit is a quantum computer's tiniest data unit. The qubit can be in primary states represented as Dirac notation using $|0\rangle$, $|1\rangle$ or linear superposition states of $|0\rangle$ and $|1\rangle$.

A physical system can be in one of many arrangements of particles or fields. According to the quantum superposition principle, the state combines all these possibilities. The qubit representation is probabilistic and is defined as a pair of numbers $(\alpha, \beta)$ for two possible arrangements, 0 and 1, of particles. Equation (1) describes the physical system as a qubit state [31].

$$|\psi\rangle = \alpha \, |0\rangle + \beta \, |1\rangle \quad \ldots (1)$$

where $|0\rangle$ and $|1\rangle$ are two basic states, the coefficients $\alpha$, $\beta$ are complex numbers with $|\alpha^2| + |\beta^2| = 1$ and dictates the probabilities of the system to be in either arrangement. $|\alpha|^2$ indicates the probability of the qubit being at state 0, and $|\beta|^2$ is the probability of the qubit being at state 1. The two basic states $|0\rangle$ and $|1\rangle$ are called computing base states of quantum bits, and they correlate to the two states 0 and 1 of digital computer bits. The notable difference between digital computer bits and qubits is that the qubits can be in a superposition state of $|0\rangle$ and $|1\rangle$ as presented in equation (1).

Quantum entanglement presents a physical phenomenon of a system where a quantum state must not be factored as a product of states since the individual component is incomplete and cannot be described without considering the other features. Moreover, they are not individual particles but are inseparable wholes. The composite system state can always be expressed as a sum or superposition [38]. In quantum computer science and information processing, entanglement is a valuable physical resource and a prominent feature of multiple qubit systems. It is simple to realize that plenty $(x + y)$ qubit states couldn't be formulated as the tensor product of an x qubit state and an y qubit state as these are entangled states. For example, the entangled states that are maximally entangled (Bell states) and weakly entangled states are the same as $|00\rangle + 0.01 |11\rangle$, also the separable ones such as $|00\rangle$.

Another concept and principle of quantum physics is interference, in which particles can be in more than one place at any given time using superposition and cross their trajectory to interfere with the direction. To realize the quantum interference, examine a generic superposition for n-qubits $\sum_n z_n |n\rangle$. The direct measure of $z_n$, only returns the local information for the possible values of n. However, the return results get changed once a unitary transformation is performed using:

$P|n\rangle = \sum_y p_{ny} |y\rangle$ for all n

on it, and then,

$$P\left(\sum_n z_n |n\rangle\right) = \sum_n z_n \left(\sum_y p_{ny} |y\rangle\right)$$

$$= \sum_y \left(\sum_n z_n p_{ny}\right) |y\rangle$$

If now we are measuring $P(\sum_n z_n |n\rangle)$, global information for all $z_n$ can obtain through amplitude $\sum_n z_n p_{ny}$ for a single value of y.

## B. Logistic Chaotic Map

The chaotic logistic map is optimal from the tenfold adapted chaotic maps in [39]. We embedded chaos theory in the search strategy of SSA. We obtained the superior average value of the optimal solution with better balance corresponding to the original SSA and another meta-heuristic algorithm.

Chaos theory is a prominent mathematical strategy and has been used extensively in the literature to improve the performance of meta-heuristic algorithms. It is typically outlined as the simulator of the dynamic behavior of a non-linear scheme. A chaotic logistic map is employed with a genetic algorithm for encrypting the image. It is utilized to encrypt the early version of the image, and later GA is executed to boost the encryption results. A hybrid global optimization algorithm is proposed based on a chaos search strategy with a complex method to jump out from the local optima obtained.

We used the following equation in this article for the chaotic logistic map. The output of this equation is embedded into the equation of QCSSO, which prevents the problem of search stagnation:

$$w_{t+1} = d * w_t (1 - w_t) \dots (2)$$

where $w_t$ is the value of the chaotic map at the t-iteration. In this article, the initial condition of the chaotic map is set to ($w_0 = 0.70$).

## C. The QCSSO Algorithm

SSA simulates the swarming performance of Salp throughout the optimization and modeling of a Salp chain. This chain may prevent stagnancy in the local optima up to a bit level because Salp is usually attracted towards the global optima by collaborating with leaders Salp. Also, the original SSA search process is ineffective in strengthening the obtained global best, i.e., to achieve global optima in the fitness landscape. The initial search process considers the optimization problem as a single problem instance. For DOPs, it is necessary to consider dynamic changes, respectively, such as the emergence of new cases of problems that must be tackled from scratch. Consequently, the standard SSA does not have the competence to examine and preserve the diversity of the population. Hence, it cannot solve dynamic optimization problems, where it is necessary to discover and monitor the global optima in dynamic and uncertain environments.

In this paper, we extended the work on a quantum-inspired algorithm based on SSA with a multi-population mechanism to discover and follow the global optima in the search space, an effective method for dynamic optimization problems (DOPs) [10]. The basic procedure and strategy of QCSSO are explained as follows:

- Multiple peaks exist in almost all dynamic environments, and each can transform into a global optimum when the environment changes. That means each peak can be a probable optimum. Hence, an optimization algorithm intended for DOPs should quickly track all the environmental peaks to identify the optimum peak on environmental change. The multi-population mechanism is a promising strategy that enhances the coverage of potential multiple optima in the search space [13].

- In the proposed multi-population strategy, a part of the population is detached from the total population to create a new sub-population. No information is being shared between the populations, excluding the duplication search among the two best individuals of two subpopulations to avoid multiple local best at a certain distance.

An approach to extend and change the original SSA with Quantum computing, multi-population mechanism, and the introduced chaotic operator could be a promising solution for the dynamic optimization problems. It accelerates the speed of SSA and supports locating and tracking the global optima, increasing the diversity of individuals and preventing stagnation in local optima in the search space.

In quantum science, the Delta potential well model describes using the Dirac delta function, which is a general function and objectively correlates to the potential that is zero universally except for a single point when it takes an infinite value. In the QCSSO, the Salp in the Delta potential well should move in the bound state in line with the strategy proposed for the particles in [40]. The position of Salp is necessary to be measured to assess the fitness of each Salp. However, only the probability of position for each Salp ($X^i$) can be find-out from the probability density function $|\Psi(x,t)|^2$, i.e., Salp emerges at position $x$ relative to point A. Thus, it is necessary to measure Salp's position thanks to the collapsing methodology, i.e., transforming from a quantum state to a classical form. This measurement process can be simulated using Monte Carlo Method using the procedure mentioned in [40].

In this article, we use the following iterative equation for the position to measure each Salp according to the proposed equation in the quantum-inspired PSO for particles in [41].

$$X_{j+1}^k = \begin{cases} A_d + B_l * |BestMean_l - X_j^k| * \ln(^{r_d}/_{u_d}) & c_3 > 0 \\ A_d - B_l * |BestMean_l - X_j^k| * \ln(^{r_d}/_{u_d}) & c_3 < 0 \end{cases}$$
$$\dots (3)$$

$$X_j^i = \left( X_j^{i-1} + C_i^{l+1} \left( A_j - X_j^i \right) + m * (X_j^{i-1} - X_j^{i-2}) \right) \dots (4)$$

$$u_d = 3 * w * (1 - w) * c_4$$

where $r_d, c_3, c_4$ are uniformly generated random values inside $[0,1]$, and $u_d$ is chaotic operator equation, $x_j^i$ indicates the position of $i^{th}$ Salp in $j^{th}$ dimension, $A_j$ reveals the local attractor for the convergence speed in the search space $j^{th}$ dimension, $B_l$ is known as the contraction-expansion coefficient, which gradually decreases through iterations. The coefficient $C_i^{l+1}$ is the most important parameter in the QCSSO, as presented in equation (5) for the follower Salp to support better coordination during the exploitation of the search space.

$$C_i^{l+1} = 0.75 * \sin\left(\pi/4\right) * \left(1 - \left(l/L\right)\right) \dots(5)$$

as l specifies the present iteration, L is the maximum number of iterations. We assumed here the first iteration $l = 0$ with maximum iteration size L. $B_l$ describes as equation (6), a contraction-expansion coefficient that gradually decreases or increases iteration-wise in line with the progress of respective Salp's convergence speed and execution of the algorithm. $A_d$ as in equation (7), a base point for Salp to move around in the vicinity. In addition, to know as an inclining learning point for Salp to oscillate around. $BestMean_l$ describes mathematically as equation (8), the mean of the individual best position. $X_j^k$ is the k-th Salp in j-dimension and $X_{j+1}^k$ is the new position of Salp.

$$B_l = \left(\frac{0.5 * (L - l)}{l + 0.5}\right) \dots (6)$$

where l is the current iteration and L is the maximum number of iterations.

$$A_d = \left(r_{1d} * X_j^k + r_{2d} * F_j\right) / r_{1d} + r_{1d} \dots (7)$$

where $r_{1d}$ and $r_{2d}$ are uniformly distributed random numbers in the range [0, 1]. $F_j$ is the food position as the best location.

$$BestMean_l = \frac{1}{N} \sum_{j=1}^d x_{k,j}(l) \dots (8)$$

where N is the maximum number of the population.

### D. Methodology

A few concepts are suggested in papers to enhance the efficiency of SSA. Still, we employed the standard SSA and implemented quantum theory, multi-population with a chaotic logistic operator as explained in the previous section, to boost the obtained optimal global solution by SSA and monitor the trajectory of the global optima. Adapting the chaotic logistic operator supports preserving individuals' diversity and precluding re-initializing the population when the change is determined, as it offers an acute information depletion. Maintenance of diversity is essential for dynamic optimization as the global optimization of DOPs evolves through time. If the Salp is grouped in a narrow area, then the individual Salp cannot discover the changes in the problem. Also, the explorative ability of EAs is dependent and determined on the population diversity, which means the exploring ability is reduced in the identical densities of the population.

In QCSSO, equation (3) and (4) updates the Salp positions and produces new solutions. The population is divided into subpopulations. An overlapping search technique between the two local best is employed to avoid entrapment in local optima and increase the searchability of the algorithm. The value of parameters $C_i^{l+1}$ makes adjustments through the loop of the algorithm and produces new adapting values for the position to update throughout the optimization process. The quantities $r_k$ k $\in \{1, 2, 3\}$ stand for evenly distributed random values within the range [0, 1]. The value of $w$ was taken as a fixed value of 0.96. Pseudo-code for the QCSSO presented in Algorithm 1 first initializes the population, the best position (i.e., food position), and other algorithm parameters. Next, the Salp chain is built using the standard equation of SSA [1]. The Salp moves towards the optimal solutions using the quantum-inspired equations of QCSSO (3) and (4), followed by the chaotic logistic mechanism to locate and track the global optima. A chaotic logistic map approach was applied while improving the position of Salp to maintain diversity. Since the nature of the SSA algorithm is iterative, it repetitively produces and develops some random Salp within the maximum and minimum limits of search space. Then, all the Salp, i.e., leader and followers, update their position in the location vector during optimization. A flow diagram of the QCSSO optimization algorithm, as displayed in Fig.2.

### Algorithm 1 Pseudo Code of QCSSO

| | |
|---|---|
| 1. | **method** QCSSO (M, N, l, u, dim, rObj) |
| 2. | **INIT:** P, R, fbest, ifitness, subP |
| 3. | **WHILE** N ≤ M **DO**　　　*Main Loop |
| 4. | 　　Multi-population Strategy |
| 5. | 　　**if** N ≥ 1 **THEN** |
| 6. | 　　　**Positions Update equation (3), (4)** |
| 7. | 　　　Values of Parameters update |
| 8. | 　　　Fitness Calculation |
| 9. | 　　**else** |
| 10. | 　　　Salp chain create using standard SSA [1] |
| 11. | 　　Evaluate local best in the subP |
| 12. | 　**end** WHILE |
| 13. | Best Solution　　　*Returns the Best Solution |

### E. Considerations for DOPs in SSA

To address the key issues of SSA for DOPs, such as quick convergence, and to maintain diversity in solution and population, the QCSSO approach with multi-population chaotic theory employ to locate and track the multiple local optima in search space. Multi-population is an effective and popular strategy used in the literature to facilitate the performance of evolutionary algorithms for dynamic optimization problems. Multi-population, where the entire population is divided into several subpopulations of small sizes to discover and monitor the constantly evolving global optima in a dynamic and uncertain environment. This strategy also helps control the solution's diversity and convert local optima into global best, which contributes to making the SSA search strategy more effective for DOPs. The initial assumption of local optima is considered in sub-population, and an overlapping search between the two best solutions is performed to avoid multiple convergences around the same point. No other information is shared between the populations. The superposition searching is performed among two subpopulations by comparing the best Salp in subpopulations. The attempt is to generate a new point around the best Salp in the subpopulations by adding a random value sampled from the normal distribution in the dimension loop.

### Algorithm 2 Ageing approach

1. In the population loop
2. **if** i-th Salp is global best then do nothing
3. **else if** i-th Salp is best in sub-population and bAge is greater than maxLimit and rRand less than 0.1
4.     re-initialize i-th Salp
5. **else if** i-th Salp is not best but bAge is greater than minLimit and rRand less than 0.1
6.     re-initialize i-th Salp
7. **until** max-pop size

Another mechanism to check the age of each Salp based on the number of generations has been considered for the Salp stagnates into local optima, which is an addition to the chaotic logistic map. The idea is to look for the Salp stagnates in local optima and re-initialize it for the next generation. According to the aging strategy, the global best is not considered. Only the local best whose age is more significant than a defined limit and no improvements during the specified number of iterations is considered to re-initialize as presented in Algorithm 2. The age of individual Salp increments by 1 in each iteration, and its fitness is evaluated to see improvement. In case of no improvements for several iterations, a flag is triggered to re-initialize it for the next generation.

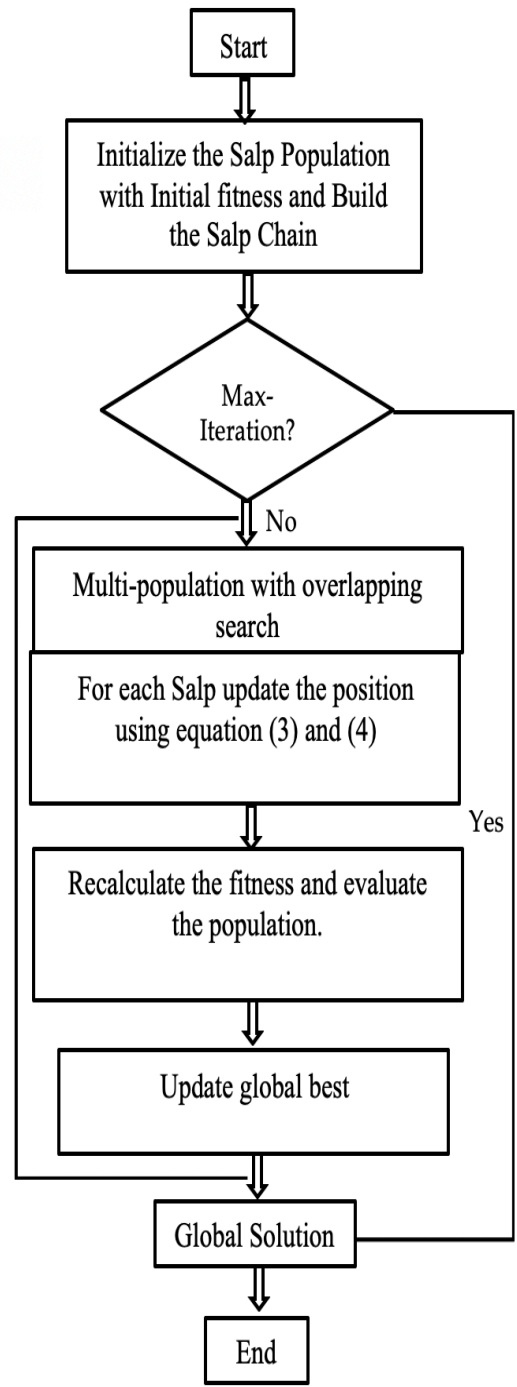

Fig. 2. Flowchart of QCSSO Optimization Algorithm.

| Function Name | Function | DIM | Range | $f_{min}$ |
|---|---|---|---|---|
| Sphere's Function | $F(x) = \sum_{i=1}^{n} x_i^2$ | 10 | [-100, 100] | 0 |
| Rastrigrin Function | $F(x) = \sum_{i=1}^{n} [x_i^2 - 10\cos(2\pi x_i) + 10]$ | 10 | [-5, 5] | 0 |
| Weierstrass Function | $F(x) = \sum_{i=1}^{n} \left( \sum_{k=0}^{k_{max}} [a^k \cos(2\pi b^k (x_i + 0.5))] \right) - n \sum_{k=0}^{k_{max}} [a^k \cos(\pi b^k)]$ | 10 | [-0.5, 0.5] | 0 |
| Griewank Function | $F(x) = \frac{1}{4000} \sum_{i=1}^{n} (x_i)^2 - \prod_{i=1}^{n} \cos\left(\frac{x_i}{\sqrt{i}}\right) + 1$ | 10 | [-100, 100] | 0 |
| Ackley's Function | $F(x) = -20 \exp\left(-0.2\sqrt{\frac{1}{n}\sum_{i=1}^{n} x_i^2}\right) - \exp\left(\frac{1}{n}\sum_{i=1}^{n}\cos(2\pi x_i)\right) + 20 + e$ | 10 | [-32, 32] | 0 |

## IV. EXPERIMENTS

### A. Experimental Context

A personal computer environment was prepared for the empirical study. The configuration is based on the following:

- Intel® Core™ i7-3520M CPU @ 2.90 GHz
- Memory 16 GB
- Operation System Microsoft Windows 10 Home
- QCSSO's code is developed in CPP
- GDBG Dynamic Framework in CPP

### B. Basic Test Functions

The achievement of the QCSSO optimization algorithm is assessed on the six standard functions (F) contemplated by Li et al. for the CEC'09 Special Session and Competition on "Evolutionary Computation in Dynamic and Uncertain Environments." It's a generalized dynamic benchmark generator (GDBG), a dynamic simulation framework for this study, and includes six benchmark functions [4]. The characteristics of GDBG include Seven change types for the control parameters -large and small step change, recurrent and recurrent with noise and dimensional shift, and random and chaotic change.

1. Peak Rotation Function (F1),
2. Sphere's Composition Function (F2),
3. Rastrigin's Composition Function (F3),
4. Griewank's Composition Function (F4),
5. Ackley's Composition Function (F5),
6. Hybrid Composition Function (F6)

Dynamic Framework Change Types:

1. Small Displacement (T1),
2. Large Displacement (T2),
3. Gaussian Displacement (T3),
4. Logistic Function (T4),
5. A Periodic Displacement (T5),
6. A Periodic Displacement with Noise (T6),
7. Random with Changed Dimension (T7)

All the benchmark functions with their range are presented in Table 1

### C. Method to Set Parameters and Testing

The QCSSO is designed to work with five parameters on the GDBG framework: M-Maximum Iteration, N-Current Iteration, Global Optima, l-Lower bound, u-upper bound, and rObj-Objective function reference. Rest all necessary parameters for the algorithms administered by another function of GDBG: the number of Salp and no. of dimension. In this section, the achievement of QCSSO is contrast with SPSO [1], CPSO [14], and QSSA [37].

The dynamic variant of PSO algorithms is assessed on the dynamic optimization problems and submitted in CEC'2009 for the maximum population size of 50. We have also set the population size as 50 for the QCSSO while using the GDBG framework. For the problem function $F_1$, two test numbers of peaks, i.e., 10 and 50, are being used, and for the other problem functions $F_2$ to $F_6$, one test number of peaks, i.e., 10, was used.

| | Errors | T1 | T2 | T3 | T4 | T5 | T6 | T7 |
|---|---|---|---|---|---|---|---|---|
| **QCSSO** | Avg. Best | 0.00E+00 | 0.00E+00 | 0.00E+00 | 0.00E+00 | 0.00E+00 | 0.00E+00 | 0.00E+00 |
| | Avg.Worst | 2.01E-02 | 3.23E+01 | 3.24E+01 | 6.04E-09 | 2.52E+01 | 6.16E+01 | 2.90E+01 |
| | Avg.Mean | 3.36E-04 | 2.52E+00 | 5.02E+00 | 3.19E-10 | 2.09E+00 | 2.85E+00 | 4.42E+00 |
| | STD. | 3.68E-03 | 6.86E+00 | 9.65E+00 | 1.50E-09 | 4.65E+01 | 1.04E+01 | 7.82E+00 |
| QSSA | Avg. Best | 0.00E+00 | 0.00E+00 | 0.00E+00 | 0.00E+00 | 0.00E+00 | 0.00E+00 | 0.00E+00 |
| | Avg.Worst | 2.55E+01 | 5.03E+01 | 4.74E+01 | 7.03E+01 | 4.00E+01 | 7.61E+01 | 3.43E+01 |
| | Avg.Mean | 4.04E+00 | 8.56E+00 | 1.68E+01 | 1.70E+01 | 7.09E+00 | 2.87E+01 | 3.80E+00 |
| | STD. | 6.25E+00 | 1.20E+01 | 1.74E+01 | 1.98E+01 | 8.64E+00 | 2.48E+01 | 7.11E+00 |
| SPSO | Avg. Best | 0.00E+00 | 0.00E+00 | 0.00E+00 | 0.00E+00 | 0.00E+00 | 0.00E+00 | 0.00E+00 |
| | Avg.Worst | 2.50E+01 | 4.89E+01 | 4.81E+01 | 7.55E+01 | 2.20E+01 | 8.04E+01 | 4.17E+01 |
| | Avg.Mean | 4.88E+00 | 1.07E+01 | 1.66E+01 | 2.57E+01 | 4.60E+00 | 2.88E+01 | 1.43E+01 |
| | STD. | 6.74E+00 | 1.31E+01 | 1.78E+01 | 2.06E+01 | 5.78E+00 | 2.59E+01 | 1.37E+01 |

| | Errors | T1 | T2 | T3 | T4 | T5 | T6 | T7 |
|---|---|---|---|---|---|---|---|---|
| **QCSSO** | Avg. Best | 0.00E+00 | 0.00E+00 | 0.00E+00 | 0.00E+00 | 0.00E+00 | 0.00E+00 | 0.00E+00 |
| | Avg.Worst | 5.81E+00 | 2.97E+01 | 4.01E+01 | 8.32E-01 | 8.38E+00 | 4.51E+01 | 2.80E+01 |
| | Avg.Mean | 6.07E-01 | 4.65E+00 | 1.39E+01 | 5.56E-02 | 1.17E+00 | 2.28E+00 | 2.54E+00 |
| | STD. | 1.50E+00 | 7.16E+00 | 1.11E+01 | 2.06E-01 | 1.83E+00 | 7.34E+00 | 5.08E+00 |
| QSSA | Avg. Best | 0.00E+00 | 0.00E+00 | 0.00E+00 | 1.42E-14 | 0.00E+00 | 0.00E+00 | 0.00E+00 |
| | Avg.Worst | 2.43E+01 | 4.20E+01 | 4.11E+01 | 6.13E+01 | 2.08E+01 | 7.54E+01 | 3.14E+01 |
| | Avg.Mean | 4.81E+00 | 1.16E+01 | 1.08E+01 | 1.71E+01 | 4.62E+00 | 2.93E+01 | 8.80E+00 |
| | STD. | 5.98E+00 | 1.08E+01 | 1.01E+01 | 1.59E+01 | 5.42E+00 | 2.59E+01 | 7.85E+00 |
| SPSO | Avg. Best | 0.00E+00 | 0.00E+00 | 0.00E+00 | 0.00E+00 | 0.00E+00 | 0.00E+00 | 0.00E+00 |
| | Avg.Worst | 2.97E+01 | 4.40E+01 | 4.44E+01 | 7.62E+01 | 3.28E+01 | 8.03E+01 | 4.89E+01 |
| | Avg.Mean | 7.68E+00 | 1.21E+01 | 1.70E+01 | 2.10E+01 | 5.68E+00 | 3.97E+01 | 1.37E+01 |
| | STD. | 7.22E+00 | 1.06E+01 | 1.28E+01 | 2.02E+01 | 6.57E+00 | 2.76E+01 | 1.23E+01 |

| | Errors | T1 | T2 | T3 | T4 | T5 | T6 | T7 |
|---|---|---|---|---|---|---|---|---|
| **QCSSO** | Avg. Best | 0.00E+00 | 0.00E+00 | 0.00E+00 | 0.00E+00 | 0.00E+00 | 0.00E+00 | 0.00E+00 |
| | Avg.Worst | 1.00E+01 | 5.31E+02 | 4.76E+02 | 5.78E+00 | 4.29E+02 | 2.46E+01 | 3.19E+01 |
| | Avg.Mean | 7.59E-01 | 1.76E+01 | 2.79E+01 | 3.61E-01 | 3.80E+01 | 2.71E+00 | 5.12E+00 |
| | STD. | 2.21E+00 | 6.88E+01 | 9.26E+01 | 1.15E+00 | 9.05E+01 | 6.83E+00 | 8.96E+00 |
| QSSA | Avg. Best | 7.71E-13 | 6.00E-13 | 7.74E-13 | 3.73E-13 | 0.00E+00 | 3.30E-13 | 5.63E-13 |
| | Avg.Worst | 7.42E+01 | 5.69E+02 | 5.28E+02 | 6.22E+02 | 4.94E+02 | 9.78E+01 | 4.72E+02 |
| | Avg.Mean | 2.92E+01 | 1.68E+02 | 1.00E+02 | 9.15E+01 | 1.35E+02 | 3.76E+01 | 3.82E+01 |
| | STD. | 1.99E+01 | 2.22E+02 | 1.71E+02 | 1.41E+02 | 1.73E+02 | 3.22E+01 | 6.97E+01 |
| SPSO | Avg. Best | 0.00E+00 | 0.00E+00 | 0.00E+00 | 0.00E+00 | 0.00E+00 | 0.00E+00 | 0.00E+00 |
| | Avg.Worst | 1.56E+02 | 5.97E+02 | 5.04E+02 | 3.55E+02 | 5.02E+02 | 4.97E+02 | 4.02E+02 |
| | Avg.Mean | 4.59E+01 | 1.71E+02 | 1.63E+02 | 5.36E+01 | 1.88E+02 | 7.56E+01 | 4.84E+01 |
| | STD. | 3.32E+01 | 2.15E+02 | 1.94E+02 | 6.12E+01 | 2.02E+02 | 8.31E+01 | 7.82E+01 |

The dynamic benchmark problem examination process is employed for all the benchmark problems of GDBG. An interface has been built to make all the necessary adjustments from the QCSSO execution context. The place of parameters is aligned in code to retain projected exploring and exploiting in the search space of DOPs. The QCSSO algorithm has been set to execute for the pre-defined number of assessments for the evaluation functions (F) and necessary parameters with change types (T) as input. It means no information is being shared

| | Errors | T1 | T2 | T3 | T4 | T5 | T6 | T7 |
|---|---|---|---|---|---|---|---|---|
| **QCSSO** | Avg. Best | 0.00E+00 | 4.32E-11 | 0.00E+00 | 1.45E-13 | 5.25E-01 | 0.00E+00 | 0.00E+00 |
| | Avg.Worst | 3.80E+02 | 9.30E+02 | 9.20E+02 | 1.09E+03 | 8.69E+02 | 1.11E+03 | 8.61E+02 |
| | Avg.Mean | 3.53E+01 | 6.19E+02 | 4.96E+02 | 1.19E+02 | 4.84E+02 | 2.89E+02 | 2.03E+02 |
| | STD. | 9.95E+01 | 3.65E+02 | 3.84E+02 | 3.03E+02 | 3.79E+02 | 3.98E+02 | 3.22E+02 |
| QSSA | Avg. Best | 0.00E+00 | 1.89E+01 | 5.89E+00 | 6.26E-13 | 9.90E+00 | 7.78E-13 | 1.32E-12 |
| | Avg.Worst | 8.93E+02 | 1.06E+03 | 9.84E+02 | 1.34E+03 | 9.84E+02 | 1.42E+03 | 1.00E+03 |
| | Avg.Mean | 2.78E+02 | 8.60E+02 | 7.80E+02 | 5.95E+02 | 7.96E+02 | 6.70E+02 | 6.45E+02 |
| | STD. | 3.31E+02 | 1.84E+02 | 2.40E+02 | 4.42E+02 | 2.26E+02 | 4.17E+02 | 3.50E+02 |
| SPSO | Avg. Best | 5.34E+00 | 2.99E+01 | 3.67E+02 | 5.66E+00 | 2.35E+01 | 4.36E+01 | 4.80E+00 |
| | Avg.Worst | 9.25E+02 | 1.13E+03 | 1.04E+03 | 1.37E+03 | 1.05E+03 | 1.66E+03 | 1.04E+03 |
| | Avg.Mean | 6.45E+02 | 9.37E+02 | 8.84E+02 | 7.81E+02 | 8.95E+02 | 9.43E+02 | 8.34E+02 |
| | STD. | 2.62E+02 | 1.35E+02 | 1.36E+02 | 3.02E+02 | 1.52E+02 | 3.25E+02 | 2.29E+02 |

| | Errors | T1 | T2 | T3 | T4 | T5 | T6 | T7 |
|---|---|---|---|---|---|---|---|---|
| **QCSSO** | Avg. Best | 0.00E+00 | 0.00E+00 | 0.00E+00 | 0.00E+00 | 0.00E+00 | 0.00E+00 | 0.00E+00 |
| | Avg.Worst | 2.53E+01 | 5.86E+02 | 2.89E+02 | 1.26E+01 | 5.06E+02 | 4.04E+01 | 3.25E+01 |
| | Avg.Mean | 1.90E+00 | 4.87E+01 | 4.45E+01 | 1.66E+00 | 1.10E+02 | 3.22E+00 | 5.35E+00 |
| | STD. | 4.91E+00 | 1.37E+02 | 1.31E+02 | 3.60E+00 | 1.80E+02 | 7.14E+00 | 8.38E+00 |
| QSSA | Avg. Best | 1.40E-12 | 5.01E-13 | 5.12E-13 | 8.25E-13 | 0.00E+00 | 3.38E-13 | 0.00E+00 |
| | Avg.Worst | 4.63E+02 | 6.68E+02 | 6.19E+02 | 3.74E+02 | 6.08E+02 | 7.25E+02 | 5.88E+02 |
| | Avg.Mean | 5.82E+01 | 3.05E+02 | 2.21E+02 | 4.29E+01 | 3.27E+02 | 5.96E+01 | 1.01E+02 |
| | STD. | 1.10E+02 | 2.71E+02 | 2.49E+02 | 6.66E+01 | 2.25E+02 | 1.04E+02 | 1.77E+02 |
| SPSO | Avg. Best | 0.00E+00 | 0.00E+00 | 0.00E+00 | 0.00E+00 | 0.00E+00 | 0.00E+00 | 0.00E+00 |
| | Avg.Worst | 4.76E+02 | 6.13E+02 | 6.48E+02 | 7.17E+02 | 6.72E+02 | 6.08E+02 | 6.00E+02 |
| | Avg.Mean | 5.93E+01 | 2.82E+02 | 2.73E+02 | 1.18E+02 | 3.63E+02 | 8.65E+01 | 1.16E+02 |
| | STD. | 1.04E+02 | 2.62E+02 | 2.58E+02 | 1.71E+02 | 2.49E+02 | 1.52E+02 | 1.82E+02 |

| | Errors | T1 | T2 | T3 | T4 | T5 | T6 | T7 |
|---|---|---|---|---|---|---|---|---|
| **QCSSO** | Avg. Best | 4.09E-14 | 4.26E-14 | 4.17E-14 | 4.09E-14 | 4.26E-14 | 4.09E-14 | 4.09E-14 |
| | Avg.Worst | 1.43E+01 | 1.16E+01 | 7.89E+00 | 6.37E-07 | 8.02E+00 | 5.45E+00 | 1.12E+01 |
| | Avg.Mean | 4.39E-01 | 7.06E-01 | 5.56E-01 | 1.37E-08 | 4.99E-01 | 2.43E-01 | 6.10E-01 |
| | STD. | 2.08E+00 | 1.99E+00 | 1.66E+00 | 9.59E-08 | 1.54E+00 | 8.86E-01 | 1.92E+00 |
| QSSA | Avg. Best | 4.26E-14 | 4.17E-14 | 4.26E-14 | 4.17E-14 | 4.26E-14 | 4.17E-14 | 4.26E-14 |
| | Avg.Worst | 9.17E+01 | 8.10E+01 | 9.70E+01 | 2.22E+02 | 5.67E+01 | 2.90E+02 | 2.74E+02 |
| | Avg.Mean | 3.53E+01 | 3.25E+01 | 2.65E+01 | 4.02E+01 | 2.80E+01 | 4.38E+01 | 3.04E+01 |
| | STD. | 2.62E+01 | 2.65E+01 | 2.32E+01 | 4.10E+01 | 1.84E+01 | 4.92E+01 | 4.29E+01 |
| SPSO | Avg. Best | 4.26E-14 | 4.17E-14 | 4.26E-14 | 4.17E-14 | 4.26E-14 | 4.09E-14 | 4.17E-14 |
| | Avg.Worst | 5.94E+02 | 5.52E+02 | 6.41E+02 | 9.55E+02 | 6.10E+02 | 9.10E+02 | 9.36E+02 |
| | Avg.Mean | 8.22E+01 | 5.41E+01 | 4.49E+01 | 8.42E+01 | 2.84E+01 | 9.88E+01 | 8.46E+01 |
| | STD. | 1.33E+02 | 8.84E+01 | 9.10E+01 | 1.43E+02 | 1.99E+01 | 1.63E+02 | 1.71E+02 |

during the execution of the QCSSO algorithm related to the problem change, including the number of peaks, dynamic, or dimension change.

TABLE 8
F6: RESULT ACHIEVED ON PEAK 10

| | Errors | T1 | T2 | T3 | T4 | T5 | T6 | T7 |
|---|---|---|---|---|---|---|---|---|
| **QCSSO** | Avg. Best | 4.09E-14 | 4.26E-14 | 4.17E-14 | 4.09E-14 | 4.26E-14 | 4.09E-14 | 4.09E-14 |
| | Avg.Worst | 1.43E+01 | 1.16E+01 | 7.89E+00 | 6.37E-07 | 8.02E+00 | 5.45E+00 | 1.12E+01 |
| | Avg.Mean | 4.39E-01 | 7.06E-01 | 5.56E-01 | 1.37E-08 | 4.99E-01 | 2.43E-01 | 6.10E-01 |
| | STD. | 2.08E+00 | 1.99E+00 | 1.66E+00 | 9.59E-08 | 1.54E+00 | 8.86E-01 | 1.92E+00 |
| QSSA | Avg. Best | 4.26E-14 | 4.17E-14 | 4.26E-14 | 4.17E-14 | 4.26E-14 | 4.17E-14 | 4.26E-14 |
| | Avg.Worst | 9.17E+01 | 8.10E+01 | 9.70E+01 | 2.22E+02 | 5.67E+01 | 2.90E+02 | 2.74E+02 |
| | Avg.Mean | 3.53E+01 | 3.25E+01 | 2.65E+01 | 4.02E+01 | 2.80E+01 | 4.38E+01 | 3.04E+01 |
| | STD. | 2.62E+01 | 2.65E+01 | 2.32E+01 | 4.10E+01 | 1.84E+01 | 4.92E+01 | 4.29E+01 |
| SPSO | Avg. Best | 4.26E-14 | 4.17E-14 | 4.26E-14 | 4.17E-14 | 4.26E-14 | 4.09E-14 | 4.17E-14 |
| | Avg.Worst | 5.94E+02 | 5.52E+02 | 6.41E+02 | 9.55E+02 | 6.10E+01 | 9.10E+02 | 9.36E+02 |
| | Avg.Mean | 8.22E+01 | 5.41E+01 | 4.49E+01 | 8.42E+01 | 2.84E+01 | 9.88E+01 | 8.46E+01 |
| | STD. | 1.33E+02 | 8.84E+01 | 9.10E+01 | 1.43E+02 | 1.99E+01 | 1.63E+02 | 1.71E+02 |

TABLE 9
PERFORMANCE SCORE OF EACH ALGORITHM ON DOPs

| | | F1 (10) | F1 (50) | F2 | F3 | F4 | F5 | F6 |
|---|---|---|---|---|---|---|---|---|
| **QCSSO** | T1 | 0.975645 | 0.955978 | 0.831956 | 0.449703 | 0.797564 | 0.844141 | 0.517966 |
| | T2 | 0.921049 | 0.892675 | 0.603084 | 0.0885542 | 0.53388 | 0.825569 | 0.603795 |
| | T3 | 0.881768 | 0.74245 | 0.588456 | 0.171561 | 0.599233 | 0.828525 | 0.685114 |
| | T4 | 0.97869 | 0.981183 | 0.898177 | 0.618185 | 0.822885 | 0.948448 | 0.622884 |
| | T5 | 0.930344 | 0.951612 | 0.576684 | 0.182611 | 0.475045 | 0.87228 | 0.69096 |
| | T6 | 0.895099 | 0.902026 | 0.7335 | 0.213375 | 0.722497 | 0.844983 | 0.548655 |
| | T7 | 0.886663 | 0.927386 | 0.706908 | 0.30559 | 0.720747 | 0.845844 | 0.610403 |
| CPSO | T1 | 0.942163 | 0.940825 | 0.727937 | 0.263052 | 0.687955 | 0.664818 | 0.556384 |
| | T2 | 0.892462 | 0.887899 | 0.574953 | 0.0183243 | 0.470139 | 0.611798 | 0.439927 |
| | T3 | 0.868731 | 0.837547 | 0.580325 | 0.0375368 | 0.489701 | 0.602617 | 0.431848 |
| | T4 | 0.976683 | 0.975418 | 0.899955 | 0.276901 | 0.883281 | 0.874479 | 0.630821 |
| | T5 | 0.889075 | 0.917639 | 0.568815 | 0.0271835 | 0.462767 | 0.608712 | 0.414599 |
| | T6 | 0.881828 | 0.873017 | 0.643585 | 0.0345534 | 0.568689 | 0.538666 | 0.358982 |
| | T7 | 0.85684 | 0.829573 | 0.65473 | 0.07386 | 0.572144 | 0.588575 | 0.41351 |
| QSSA | T1 | 0.776032 | 0.768278 | 0.28525 | 0.22783 | 0.314371 | 0.308042 | 0.285125 |
| | T2 | 0.733743 | 0.71041 | 0.251848 | 0.0195408 | 0.165069 | 0.322602 | 0.258774 |
| | T3 | 0.642567 | 0.718064 | 0.318484 | 0.0515741 | 0.224831 | 0.359423 | 0.303288 |
| | T4 | 0.628939 | 0.737373 | 0.233292 | 0.0949041 | 0.288899 | 0.252561 | 0.227879 |
| | T5 | 0.737373 | 0.777055 | 0.30825 | 0.0487118 | 0.17757 | 0.421606 | 0.351408 |
| | T6 | 0.51961 | 0.520214 | 0.262468 | 0.0794769 | 0.216879 | 0.243877 | 0.231668 |
| | T7 | 0.78629 | 0.747422 | 0.322689 | 0.107792 | 0.331003 | 0.36822 | 0.332159 |
| SPSO | T1 | 0.875204 | 0.83808 | 0.273157 | 0.0566925 | 0.345212 | 0.269387 | 0.234677 |
| | T2 | 0.80101 | 0.784004 | 0.22755 | 0.0181545 | 0.237943 | 0.336208 | 0.223524 |
| | T3 | 0.707618 | 0.699526 | 0.289606 | 0.0221931 | 0.198411 | 0.39036 | 0.368879 |
| | T4 | 0.604371 | 0.669144 | 0.280544 | 0.0216652 | 0.250209 | 0.19413 | 0.227945 |
| | T5 | 0.895946 | 0.877942 | 0.335961 | 0.0316889 | 0.220481 | 0.487727 | 0.328684 |
| | T6 | 0.591665 | 0.475409 | 0.169853 | 0.0140205 | 0.265328 | 0.197332 | 0.215123 |
| | T7 | 0.726484 | 0.754176 | 0.395141 | 0.0389615 | 0.365696 | 0.34295 | 0.252973 |

| Algorithm | F1 (10) | F1 (50) | F2 | F3 | F4 | F5 | F6 | Score (%) |
|-----------|---------|---------|-----|-----|-----|-----|-----|-----------|
| **QCSSO** | 0.092605555 | 0.09066272 | 0.112875096 | 0.046265181 | 0.106358448 | 0.137468208 | 0.097831424 | **68.4067** |
| CPSO | 0.0903325 | 0.0897809 | 0.106369 | 0.016963 | 0.0946551 | 0.103043 | 0.0745976 | 57.5741 |
| QSSA | 0.06843686 | 0.069231665 | 0.044993232 | 0.014253577 | 0.038598904 | 0.051686184 | 0.045109952 | 33.2310 |
| SPSO | 0.07440205 | 0.072703335 | 0.04416236 | 0.004569337 | 0.042273152 | 0.050490656 | 0.042419536 | 33.1020 |

## D. Performance Assessment of Algorithm

The total 49 test case of the six essential test functions (i.e., F1 - F6) with seven change types (i.e., T1 – T7) is considered for the performance assessment of the optimizer. We recorded the values of errors in best case average, the mean average, worst average, and STD for each possible case, which is defined as in [1]:

$$\text{Average-best} = \sum_{i=1}^{runs} Min_{j=1}^{num-change} \frac{E_{i,j}^{last}(t)}{runs}$$

$$\text{Average-mean} = \sum_{i=1}^{runs} \sum_{j=1}^{num-change} \frac{E_{i,j}^{last}(t)}{runs*num-change}$$

$$\text{Average-worst} = \sum_{i=1}^{runs} Max_{j=1}^{num-change} \frac{E_{i,j}^{last}(t)}{runs}$$

here, $E_{i,j}^{last}(t) = | Q (y^b(t)) - Q (y * (t)) |$, i.e., to calculate for reaching max change for each change type and y * (t) is the global optimum at time t.

## V. DISCUSSION ON EXPERIMENTAL RESULTS

We conduct practical experiments with numerical evaluation to prove the theoretical claims undertaken in previous sections. Numerical results are gathered in the sum of marks obtained in each case and multiplied by 100 to measure the score in percentage. The original SPSO, QSSA, and QCSSO score is evaluated in the said environment, and reference score values of CPSO are as per the score obtained in [14]. This section presents the results of the experiments, including the comparison report with peer algorithms and analysis of different change effects for each function during the optimization process.

### A. Experimental Analysis

The QCSSO is implemented and assessed for dynamic optimization problems (DOPs). The measured value of execution on each problem is recorded in the form of best average, mean average, worst average, and standard deviation (SD), tabled in Table 2 to Table 8 as a result of the analysis. In Table 9, the score of each algorithm is lodged with all six benchmark problems and seven change types and with the combination of different test cases. Assessment results and data analysis show that QCSSO has performed better on most benchmark problems than SPSO, QSSA, and CPSO optimization algorithms.

### B. Why QCSSO Performed Better Than Standard SSA

The searching process of standard SSA is designed for stationary optimization problems in which quick convergence is considered a good feature. The search strategy of standard SSA is not appropriate for dynamic optimization problems where it is necessary to enhance the obtained global optimal solution so far to achieve the foreseen global optima. The original search strategy considers the optimization problem as a single problem instance. For DOPs, it is necessary to consider each dynamic change as a new problem case that must be addressed from scratch. Hence, a better search strategy is required for the DOPs to conform to the dynamic changes, i.e., by forwarding the experience of the optimization process, considering the new environment is somewhat related to the old one. Also, population diversity is lost in the standard SSA due to expected convergence for the stationary optimization problems. Preserving the population's diversity is crucial for dynamic optimization as the global best changes over time. If the population is collected in a tight region, the individual may not detect changes in the landscape. The obtained solution will not be improved further. From the trajectory path of standard SSA, it can be derived that all the Salp intensely move towards the guided approach by leader Salp in the direction of food position and sometimes converge on local or global optima where food position is located. Because of this feature, the standard SSA is ineffective in discovering and monitoring the evolving global optima in the search landscape.

The QCSSO employs a different mechanism to enhance the performance of standard SSA for dynamic optimization problems. First, a better search strategy is considered to use the experience during the optimization process when a change is detected along with quantum computing to tune the original algorithm for the specific instances of DOPs during the optimization process by taking the actual progress of the search. Second, multi-population with an aging mechanism is applied to discover and monitor the ever-changing global optima in the search landscape. This strategy, in addition to the chaotic logistic operator, helps to control the diversity and convert good local optima into global best during the optimization process in the changing environment.

### C. Comparative Study and The Effect of Dynamic Changes

The performance of QCSSO is measured for all the change types of the dynamic benchmark functions. QCSSO performed better than all its peer algorithms, including CPSO, SPSO, and QSSA. The employed strategy in QCSSO to locate and track global optima; outperformed and indicated its superiority over CPSO, SPSO, and QSSA. The QCSSO achieved excellent results for F1(on both peaks), F2, F5, and F6 test functions of DOPs for all the change types. However, CPSO performed close to QCSSO for chaotic (T4) change types. The hierarchical clustering method and local search strategy enabled CPSO to converge faster and have a better result for the test function F4, chaotic (T4) change types. The regional search strategy helps in searching for optimal solutions in promising sub-regions detected by the clustering method to exploit it effectively. However, the clustering approach is ineffective in generating the sub-swarms consistently, especially in the case of a single particle covering a peak, because there is no improvement during the optimization process in that cluster. From the overall final score in Table 10, the superiority of the QCSSO algorithm can be easily made out.

The algorithm has obtained good results across all change types for functions F1 (on both peaks), F2, and F5. QCSSO has steady achievement on problem F6, i.e., Hybrid composite problem for all seven change types (T1-T7), discovered by the comparative study of best obtained average values from Table 8. Benchmark problem function Rastrigin's (F3) for the change large step (T2) appears as the complex test case for the QCSSO algorithm between all the dynamic optimization benchmark functions.

*D. Results of The Experiment*

The overall performance table on DOPs of QCSSO shows the algorithm's superior capability of locating and tracking multiple optima. The algorithm has obtained excellent results across all change types for functions F1(rotation peak on both peaks), F2 (Composition of Sphere's function), and F5(Ackley's function) test functions of DOPs for all the change types when it is compared with CPSO, SPSO, and QSSA. Overall, algorithm performance is suitable across all functions. It has performed very well for functions F2, F4, and F5 when comparing the overall score. The F3 function (Composition Rastrigin function) for a large step is the most challenging problem amongst all dynamic benchmark functions of the GDBG framework for the QCSSO algorithm, but still better than the peer algorithms. This analysis is reflected in the overall performance in Table 10, where QCSSO scores highest when compared with other well-regarded algorithms.

## VI. Conclusions And Future Work

The current work proposes an extension of the Salp Swarm Algorithm (SSA) with multi-population, quantum computing, and the chaotic logistic map for the dynamic optimization benchmark problems (DOPs) presented in the CEC'2009 special session. For DOPs, an optimization algorithm usually must discover and monitor the multiple optima changing over time. The desirable features of an optimization algorithm include maintaining diversity and multi-population for locating and tracking the optima. In this article, the multi-population is used to discover and monitor the global optima, and quantum computing techniques are used in the equation to increase the searchability of the algorithm. The strategy also helps control the solution's diversity and convert good local optima into global best during the optimization process. Further, a chaotic operator is employed to maintain diversity at the individual level and avoid entrapment in local optima. Implementing chaotic series instead of random in QCSSO is a robust approach to diversify the population, enhance overall performance, and prevent early convergence.

As illustrated in the current contribution, the trial was carried out to evaluate the performance of QCSSO. The proposed algorithm is compared with well-regarded algorithms: QSSA, SPSO, and CPSO. The six dynamic benchmark optimization problems (F1-F6) evaluation results with seven change types (T1-T7) were recorded and analyzed. It shows that QCSSO markedly improves SSA's performance in locating and tracking multiple optima in a dynamic fitness landscape and can find an acceptable solution for most of the DOPs in dynamic and uncertain environments.

Although the currently applied method effectively boosts the performance of SSA for DOPs, a fixed sub-population size is considered in this work, which is not a good approach. Thus, more work can be done to make it self-adaptive based on the optimization progress.